\documentclass{llncs} %
\usepackage{graphicx}
\usepackage{subfigure}
\usepackage{array}
\usepackage{bm}
\usepackage{color}
\usepackage{soul}
\begin{document} 
\title{Subject2Vec: Generative-Discriminative Approach from a Set of Image Patches to a Vector}
\author{Sumedha Singla\inst{1}, Mingming Gong\inst{2}, Siamak Ravanbakhsh\inst{3}, Frank Sciurba\inst{4}, Barnabas Poczos\inst{5} \and Kayhan N. Batmanghelich\inst{1,2,5} }
\institute{Computer Science Department, University of Pittsburgh, USA \and Department of Biomedical Informatics, University of Pittsburgh, USA \and 
Computer Science Department, University of British Columbia, Canada \and
University of Pittsburgh School of Medicine, University of Pittsburgh, USA \and
Machine Learning Department, Carnegie Mellon University, USA }
\maketitle %

\newcommand\ie {{\it i.e., }}
\newcommand\eg {{\it e.g., }}
\newcommand\etc{{\it etc.}}
\newcommand\cf {{\it cf. }}
\newcommand\etal {{\it et al.}}
\newcommand\eq {{\it Eq.}}

\begin{abstract} 
We propose an attention-based method that aggregates local image features to a subject-level representation for predicting disease severity. In contrast to classical deep learning that requires a fixed dimensional input, our method operates on a \emph{set} of image patches; hence it can accommodate variable length input image without image resizing.  The model learns a clinically interpretable subject-level representation that is reflective of the disease severity. Our model consists of three mutually dependent modules which regulate each other: (1) a \emph{discriminative} network that learns a fixed-length representation from local features and maps them to disease severity; (2) an \emph{attention} mechanism that provides interpretability by focusing on the areas of the anatomy that contribute the most to the prediction task; and (3) a \emph{generative} network that encourages the diversity of the local latent features. The generative term ensures that the attention weights are non-degenerate while maintaining the relevance of the local regions to the disease severity. We train our model end-to-end in the context of a large-scale lung CT study of Chronic Obstructive Pulmonary Disease (COPD). Our model gives state-of-the art performance in predicting clinical measures of severity for COPD. The distribution of the attention provides the regional relevance of lung tissue to the clinical measurements.  
\end{abstract}
\section{Introduction} 
We propose a deep learning model that learns subject-level representation from a \emph{set} of local features. Our model represents the image volume as a \emph{bag} (or set) of local features (or patches) and can accommodate input images of variable sizes. We target diseases where the pathology is diffused and is not always located in the same anatomical region. The model learns by optimizing the objective function that balances two goals: (1) to build a fixed length subject-level feature that is predictive of the disease severity, (2) to extract interpretable local features that identify regions of anatomy that contribute the most to the disease. Our motivation comes from the study of COPD, but the proposed model is applicable to a wide range of heterogeneous disorders.

Many diseases such as emphysema are highly heterogeneous~\cite{Satoh2001CTSmokers} and show diffuse pattern in computed tomographic (CT) images of the lung. Having an objective way to characterize local patterns of the disease is important in diagnosis, risk prediction, and sub-typing~\cite{Shapiro2000EvolvingDisease.,Hayhurst1984DiagnosisTomography,Muller1988DensityTomography.,Estepar2013ComputedImplications}. Although various intensity and texture based feature descriptors are proposed to characterize the visual appearance of the disease~\cite{Cheplygina2017TransferDisease,Sorensen2012Texture-basedApproach,Yang2017UnsupervisedStudy}, most image features are generic and are not necessarily optimized for the disease. Recent advances in deep learning enable researchers going directly from raw image to clinical outcome without specifying radiological features~\cite{Gonzalez2017DiseaseTomography,Dubost2017Gp-Unet:Network}. However, the classical deep learning methods, that operate on entire volume or slices\cite{Gonzalez2017DiseaseTomography}, are challenging to interpret and they require resizing the input images to a fixed dimension. That is particularly the case for lung images due to the significant variation in the lung volume amongst individuals. Reshaping voxels in a CT image without adjusting for the density, changes the interpretation of the intensity values.

In this paper, we address these issues. We view each subject as a \emph{set} of image patches from the lung region. Different lung sizes result in bags with different number of elements. Previously,~\cite{Cheplygina2017TransferDisease,Schabdach2017AStudies} viewed the subjects as sets and used handcrafted image features. In contrast, the \emph{discriminative} part of our model uses deep learning approach and directly extracts features from the volumetric patches. Next, we use an attention mechanism~\cite{Bengio2015ShowTell} to adaptively weight local features and build the subject level representation, which is predictive of the disease severity. Our model is inspired by the Deep Set~\cite{Zaheer2017DeepSets}. However, our method adopts \emph{generative}  regularization to prevent the redundancy of the hidden features. Furthermore, the \emph{attention} mechanism provides interpretability by quantifying the relevance of a region to the disease. We evaluate the performance of our method on a simulated dataset as well as a large-scale COPD lung CT dataset where our method gives state-of-the-art performance in predicting the clinical measurements.

\section{Method} 
\begin{figure} 
\centerline{
\includegraphics[scale = 0.37]{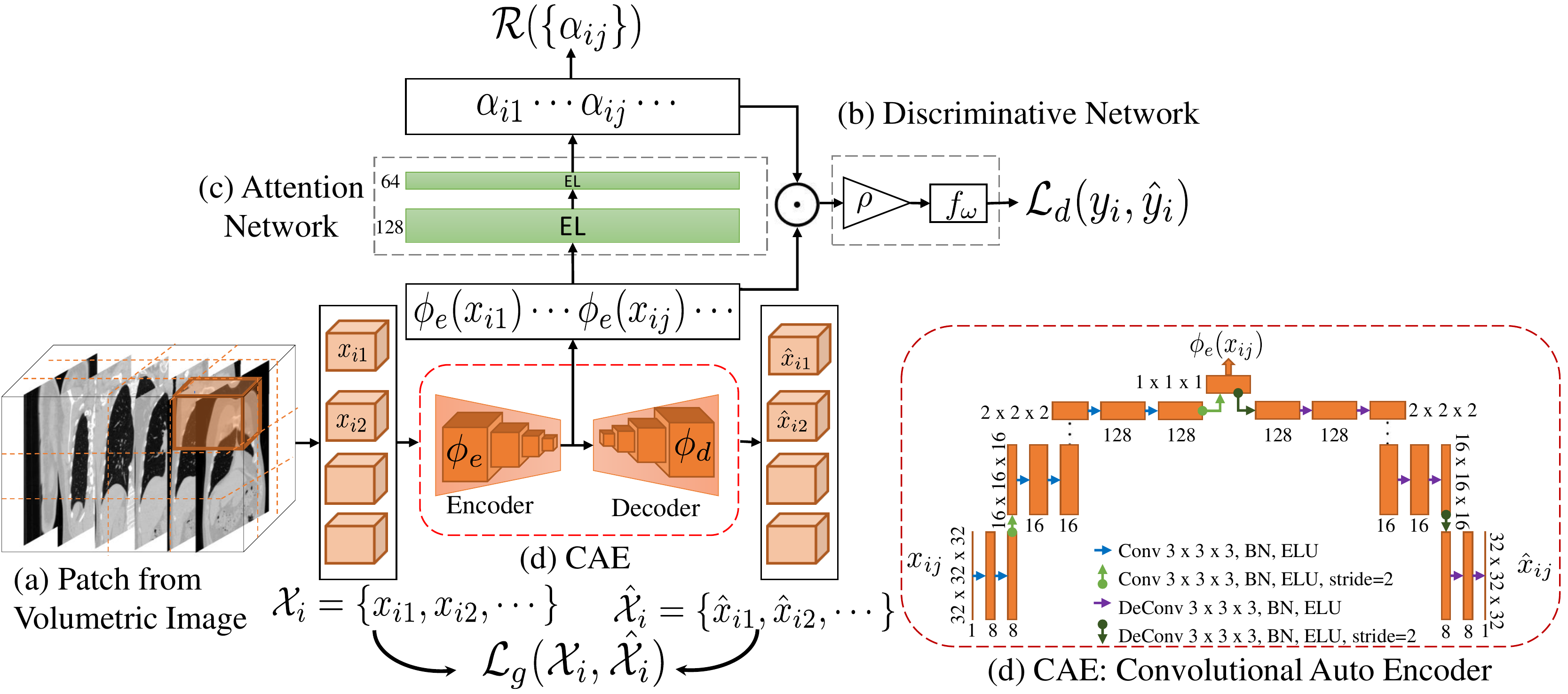}
}
\caption{(a) A subject is represented as a set of 3d image patches, (b) Discriminative Network: aggregates local features to form a fixed length representation for the subject and predicts the disease severity $(\hat{y}_i)$, (c) Attention Network: focuses attention on critical patches to provide interpretability. (d) Convolutional Auto Encoder (Generative Network): prevents redundancy of latent features.}
\label{fig:model}
\vspace{-0.5cm}
\end{figure}

We represent each subject as a set (bag) of volumetric image patches extracted from the lung region $\mathcal{X}_i = \{x_{ij}\}_{j=1}^{N_i}$, where $N_i$ is the number of patches for subject $i$. Our method maps $x_{ij}$ to a low-dimensional latent space. It then aggregates the latent features to form a fix-length representation, by adaptively weighting the patches based on their contribution in prediction of disease severity ($y_i$). The general idea of our approach is shown in Fig.\ref{fig:model}.

The method consists of three networks that are trained jointly: (1) a \emph{discriminative} network, that aggregates the local information from patches in the set $\mathcal{X}_i$ to predict the disease severity $y_i$, (2) an \emph{attention} mechanism, that helps discriminative network to selectively focus on patch-features by assigning weights to the patches in $\mathcal{X}_i$, and (3) a \emph{generative} network, that regularizes the discriminative network  to avoid redundant representation of patches in the latent space. The model is trained end to end, by minimizing the below objective function:
\begin{equation}
\label{eq:general}
\min_{\omega, \theta_e , \theta_d, \theta_a} \sum_i \mathcal{L}_d\left(y_i, \hat{y}_i ( \mathcal{X}_i ) ; \theta_e, \omega\right) +
\lambda_1  \mathcal{L}_g\left(  \mathcal{X}_i, \hat{\mathcal{X}}_i ; \theta_e , \theta_d \right) + \lambda_2 \mathcal{R} \left( \mathcal{X}_i ; \theta_e, \theta_a \right),
\end{equation}
where $\mathcal{L}_d(\cdot, \cdot)$ and  $\mathcal{L}_g(\cdot, \cdot)$ are the discriminative and generative loss functions respectively and the third term is a regularization $\mathcal{R}(\cdot)$ over the attention. The $\theta_e$, $\theta_d$, $\theta_a$ and $\omega$ are the parameters of each loss function; some of the parameters are shared. The $\lambda_1, \lambda_2$ control the balance between the terms. The sum is over number of subjects. In the following section, we discuss each term in more detail.

\subsection{Discriminative Network}
The discriminative network transforms the input set of image patches and estimates the disease severity $\hat{y}_i(\mathcal{X}_i)$ as 
\begin{equation}\label{eq:4}
\hat{y}_i(\mathcal{X}_i) = f\left( \rho \left(\phi_e \left(\mathcal{X}_i, \theta_e \right)\right), \omega \right).
\end{equation}
The transformation is composed of three functions: (1) $\phi_e(\cdot;\theta_e)$ is an encoder function parameterized by $\theta_e$. It extracts features from patches in the set $\mathcal{X}_i$ and outputs a set of features. (2) The $\rho(\cdot)$ function operates on the elements of the set and converts the variable length set $\phi_e(\mathcal{X}_i;\theta_e)$ into a fixed length vector. It is a permutation invariant function such as, maximum function $\rho(\cdot) = \max (\phi_e(x_{i,1}), \cdots , \phi_e(x_{i,N_i}))$ or mean function  $\rho(\cdot) = \frac{1}{N_i} \sum_{j=1}^{N_i} \phi_e(x_{ij})$. This formulation ensures that, $\hat{y}_i(\mathcal{X}_i)$ is invariant to the order of patches in $\mathcal{X}_i$. In our experiments, we tried different $\rho$'s and the mean function works well for our task. The mean function  assumes all the instances within the set are contributing equally to the set level feature vector. We extended it further to perform weighted mean, where weights are learned using the attention network in Section 2.2. (3) $f(\cdot;\omega)$ is a prediction function, parameterized by $\omega$. It takes the set-level feature vector extracted by $\rho(\cdot)$ as input, and estimates the disease severity. In this paper, we predict two continuous clinical variables. Finally, $\mathcal{L}_d\left(y_i, \hat{y}_i ( \mathcal{X}_i ) ; \theta_e, \omega\right)$ is a $\ell_2$ loss function between predicted and true value.



\subsection{Attention Mechanism}
The goal of our proposed model is twofold: first to provide a prediction of the disease severity and secondly, to provide a qualitative assessment of our prediction, to enhance the interpretability of the results. For the given problem, it is reasonable to assume that different regions in the lung contribute differently to the disease severity. We model this contribution by adaptively weighting the patches. The weight indicates the importance of a patch in predicting the overall disease severity of the lung. This idea is similar to the concept of attention mechanism in the Computer Vision~\cite{Bengio2015ShowTell} and Natural Language Processing~\cite{Luong2015EffectiveTranslation} communities. 

We estimate the attention weights for the subject $i$ $\left( \bm{\alpha_i} = \{ \alpha_{i1}, \cdots , \alpha_{i,N_i}\} \right)$ by the attention network as
\begin{equation}
 \bm{\alpha_i} = A\left( \phi_e \left( \mathcal{X}_i ; \theta_e\right); \theta_{a} \right).
\end{equation}
Unlike the $\rho(\cdot)$ function in Section 2.1, $A(\cdot; \theta_a)$ maps a set to another set. Permuting the order of elements in the set $\mathcal{X}_i$, should 
\emph{equivariantly} permute the output set $ \bm{\alpha_i}$. To ensure $A(\cdot)$ is a permutation equivariant function, we construct it as a neural network with equivariant layer (EL)~\cite{Zaheer2017DeepSets}. Assuming $\mathbf{H}_i \in \bbbr^{N_i, d}$ where $k^{th}$ row is $\phi(x_{ik} ; \theta_e) \in \bbbr^d$, one possible way of modeling the equivariant layer is
\begin{equation}\label{eq:EL}
\left[ \mathbf{H}_i\right]_k = \mathbf{W}\left(\left[ \mathbf{H}_i\right]_k - \max(\mathbf{H}_i, 1)\right) + \mathbf{b},
\end{equation}
where $\left[ \mathbf{H}_i \right]_k $ denotes $k^{th}$ row of $\mathbf{H}_i$ and $\max(\mathbf{H}_i,1)$ is the max over rows. $\mathbf{W} \in \bbbr^{L \times d}$, $\bm{b} \in \bbbr^L$ are the parameters of the EL. It is straightforward to show that the function in \eq~\ref{eq:EL} is permutation-equivariant. To ensure $A(\cdot; \theta_a)$ is permutation equivariant we construct it by composing few EL's. Also, we assume that the weights $(\bm{\alpha_i})$ are non-negative numbers that sums to 1. We pass the output of the EL to a softmax to obtain a distribution of weights over the patches in the subject. Finally, to ensure the weights are sparsely distributed, we added a regularization term $\mathcal{R} \left( \mathcal{X}_i ; \theta_e, \theta_a \right) = \sum_{j=1}^{N_i} \log (\alpha_{ij} + \epsilon)$ to the loss function in \eq~\ref{eq:general} 
\subsection{Generative Network}
The encoder function  $\phi_e$  projects the raw image patch $x_{ij}$ to a $d$-dimensional latent representation  $\left( \ie  \phi_e(x_{ij}; \theta_e) \in \bbbr^d \right)$. Without extra regularization, the loss function focuses only on the prediction task, forcing the encoder function to extract information that is only relevant to $y$. If $y$ is low dimensional, the encoder function learns a highly redundant latent space representation for each patch. Since $\alpha_{ij}$ is a function of $\phi_e(x_{ij}, \theta_e)$, redundant features result in almost uniform weights \ie ($\alpha_{ij} = \frac{1}{|\mathcal{X}_i|}$). This phenomenon makes interpretability very difficult. We demonstrated this effect in our experiments. 

To discourage this loss of information, we added a convolutional auto-decoder (CAE)~\cite{Masci2011StackedExtraction} to reconstruct input patch as $\hat{x}_{ij}= \phi_d(\phi_e(x_{ij}; \theta_e); \theta_d)$. Finally, we add a generative loss $\mathcal{L}_g(  \mathcal{X}_i, \hat{\mathcal{X}}_i ; \theta_e , \theta_d ) = \frac{1}{\left|\mathcal{X}_i \right|}\sum_{x_{ij}\in \mathcal{X}_i} || x_{ij} - \hat{x}_{ij} ||_2$ to the cost function in \eq~\ref{eq:general}.

\subsection{Architecture Details}
The $f(\cdot;\omega)$ is a linear function predicting the disease severity $y_i$. The architecture of generative network is elaborated in Fig.\ref{fig:model}. The convolutional layer employs batch-normalization for regularization, followed by an exponential linear unit (ELU)~\cite{Clevert2015FastELUs} for non-linearity. The attention network $A(\cdot; \theta_a)$ has 2 equivalence layers with sigmoid activation function, followed by a softmax layer. The model is trained using Adam optimizer~\cite{Kingma2014Adam:Optimization} with a fixed learning rate of 0.001.

\section{Experiments}
We evaluate the prediction and interpretation of our method on synthetic and real datasets. To evaluate the interpretability of our method quantitatively, we synthesize a dataset where the set-level target ($y$) are simulated from a subset of instances. Hence by viewing the attention weights as a detector of the relevant instances, we are able to evaluate the interpretability of our approach.
\begin{figure}[t]
\centering
\subfigure[ ]{\label{fig:a}\includegraphics[height=29mm]{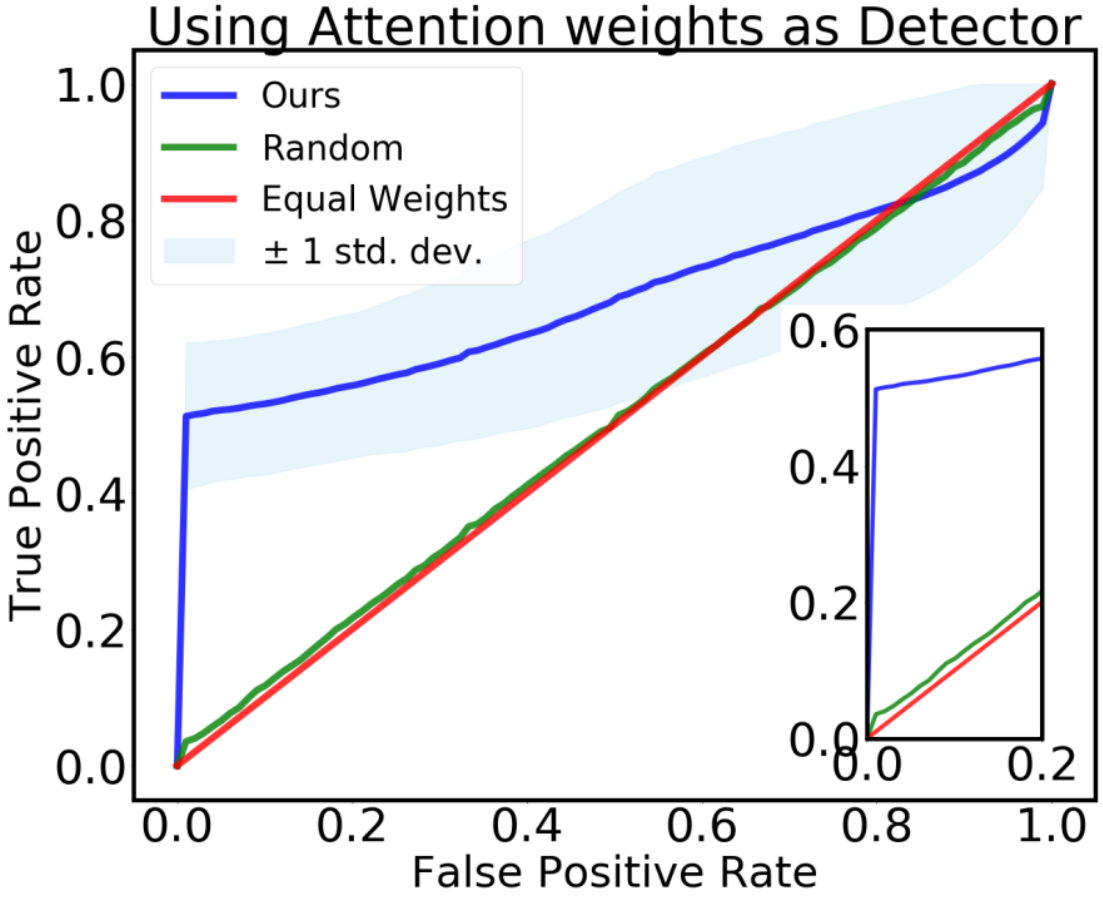}}
\subfigure[ ]{\label{fig:b}\includegraphics[height=29mm]{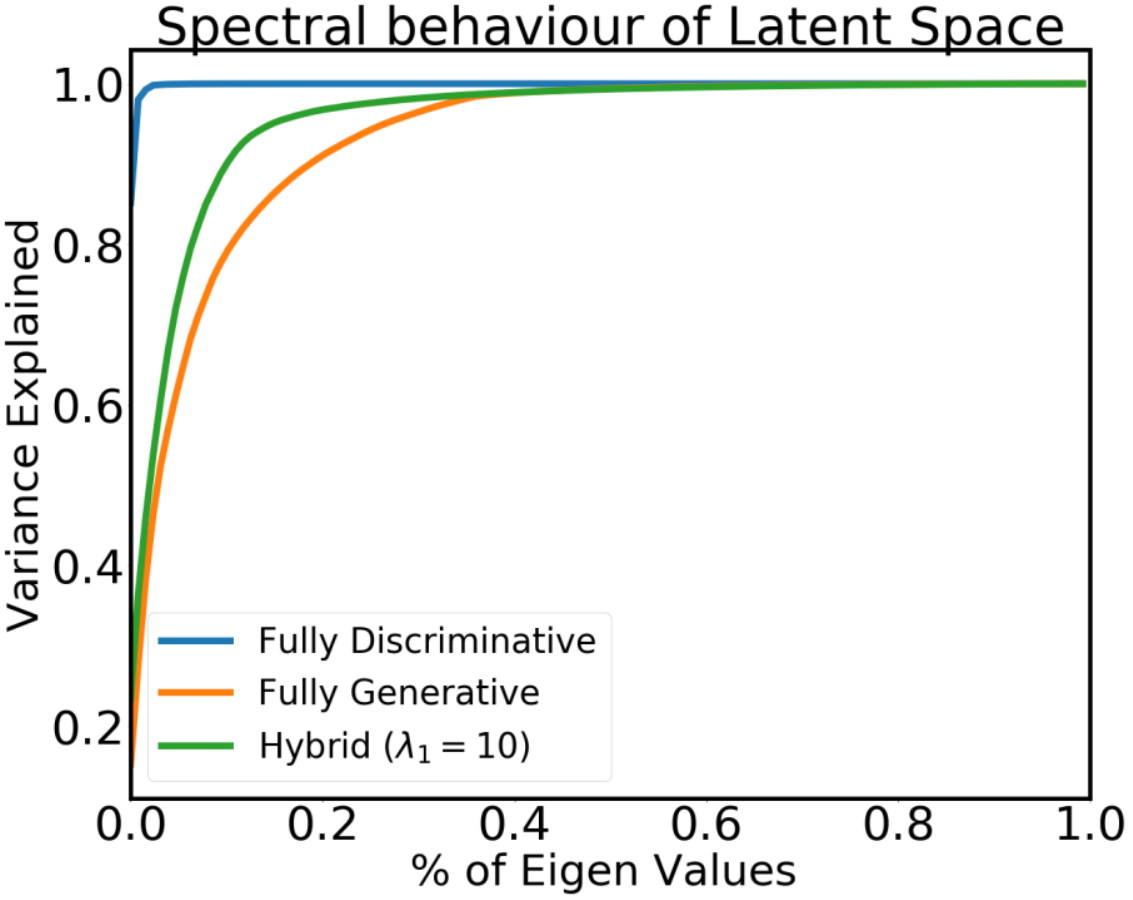}}
\subfigure[ ]{\label{fig:b}\includegraphics[height=29mm]{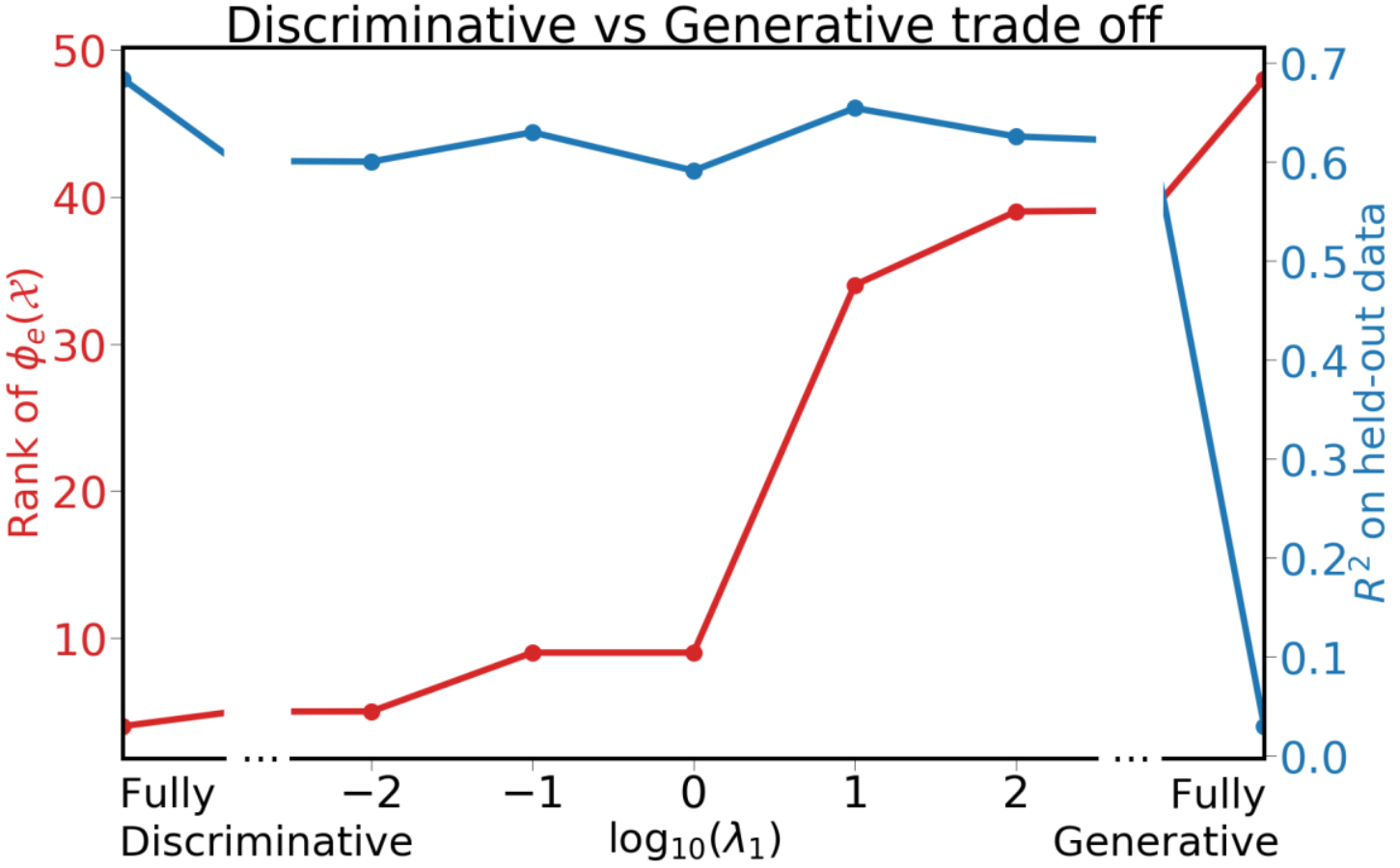}}
\caption{(a) ROC curve of detecting true relevant instances on synthetic dataset using attention weights, (b) Spectral properties of patch-level features for different values of $\lambda_1$. (c) The trade-off between rank of latent space (red, $y$-axis on left) and predictive power (blue, $y$-axis on right) for different values of $\lambda_1$. Left represents fully discriminative and right represents fully generative models.}
\label{fig:results}
\vspace{-0.3cm}
\end{figure}
\subsection{Synthetic Data}
In this experiment, we build 10,000 training and 8,000 testing sets. The instances in the set are randomly drawn images from MNIST~\cite{LeCun2010MNISTDatabase} dataset. The size of the sets varies between 20 to 100 instances. Each image is a $28\times 28$ pixel monochrome image of a handwritten digit between $0-9$. The set-label ($y$) is the sum of prime numbers ($2,3,5,7$) in that set. Our method predicts the set-label with a high accuracy ($R^2 = 0.99$ on held-out data).
We view the attention weights as detectors of prime numbers. Note that no instance level supervision is used.
 We make an ROC (Receiver Operating Characteristic) curve per set, and compute one average ROC curve across the held-out dataset. 
 Fig.\ref{fig:results}(a) shows the average and error bar for all the sets. The figure compares our method (blue) with equal weights (red) (\ie $\alpha_{ij} = 1/| \mathcal{X}_i|$) and uniform random weights (green). Our method can detect correct instances in the set, with only weak supervision over the set (\ie set-level label $y$). Here we used $\lambda_1 = 100$ and $\lambda_2 = 0.01$.

\subsection{COPD}
We evaluate our model on 6,400 subjects with different degrees of severity of the COPD from the COPDGene dataset~\cite{Regan2010GeneticDesign}. As clinical measures of the disease severity, we use the Forced Expiratory Volume in one second (FEV1), the ratio of FEV1 and Forced Vital Capacity (FVC), and discrete score (between 0-4) called the Global Initiative for Chronic Obstructive Lung Disease (GOLD). We first segment the lung area on the inspiratory images using CIP library~\cite{ross2015chest}. Each subject is represented as a bag of equal size 3D patches, with some overlap. Large patch size and percentage overlap leads to GPU memory issues. We experimented with different values and finally used patch-size of $32 \times 32 \times 32$ with 40\% overlap in our experiments. 

We perform three experiments: (1) \textit{Prediction:} we compare the performance of our method against the sate-of-art for predicting the clinical measurements, (2) \textit{Generative regularizer ($\lambda_1$):} we study the effect of the generative regularizer (\ie $\lambda_1$) in terms of prediction accuracy and information preserved in latent space, (3) \textit{Visualization:} we visualize the interpretation of the model on the subject and population level. 
Unlike $\lambda_1$, the choice of  $\lambda_2$ don't   have any significant effect on the prediction accuracy. The value of $\lambda_2$ influences the sparsity and diversity of the attention weights. In the experiments, we fixed $\lambda_2$ to 0.0001.
\begin{table}[t]
\centering
\begin{tabular}{>{\raggedright\arraybackslash}p{3.5cm}|>{\centering\arraybackslash}p{1.3cm}| >{\centering\arraybackslash}p{2.0cm} | >{\centering\arraybackslash}p{2.5cm}|>{\centering\arraybackslash}p{2.5cm}} 
\textbf{Method} & \textbf{FEV1} & \textbf{FEV1/FVC} & \textbf{GOLD exact} & \textbf{GOLD one-off}\\  
\hline
Our method ($\lambda_1=0$) &  \textbf{0.68} &  \textbf{0.71} &  \textbf{61.17} \% & \textbf{87.64} \%\\ 
Our method ($\lambda_1=10$) & \textbf{0.64} &  0.70 &  \textbf{55.60} \%& \textbf{84.57}\%\\
\hline
CNN \cite{Gonzalez2017DiseaseTomography} &  0.53  &  {---} &  51.1 \% & 74.9 \%\\ 
Non-Parametric~\cite{Schabdach2017AStudies} &  0.58  &   0.70&  50.47 \%& {---}\\ 
K-Means~\cite{Schabdach2017AStudies} &  0.54  &   0.67&   48.23 \%&{---}\\ 
Baseline&   0.52&  0.69&   49.06 \%&{---}\\ 
\hline
\end{tabular}
\label{table:1}
\caption{Clinical measurement regression and GOLD stage classification accuracy by different methods on the COPDGene dataset.
} 
\vspace{-0.5cm}
\end{table}

\textit{Prediction:} We compare to several baselines: 
(a) {\bf Baseline}: two threshold-based features measuring the percentage of voxels with intensity values less than a threshold in the images; −950 Hounsfield Unit (HU) for the inspiratory and -856 HU for expiratory. Those measurements reflect what is clinically used to quantify emphysema and the degree of gas trapping. 
(b) {\bf Non-parametric}: Schabdach et. al~\cite{Schabdach2017AStudies} view each subject as a set of hand-crafted histogram and texture features from supervoxels. They represent each subject in an embedding space using a non-parametric distance between sets. 
(c) {\bf CNN}: Gonzalez et. al~\cite{Gonzalez2017DiseaseTomography} use deep features learned from a composite image of four canonical views of a CT scan to quantify FEV1 and stage COPD. 
(d) {\bf BOW}: This method views each subject as a set of hand-crafted features from super-voxels but applies $k-$means to extract the subject-level representation.
We perform 10-fold cross-validation and report $R^2$ for the continuous measurements (\ie FEV1 and FEV/FVC) and accuracy for the GOLD score. Since the GOLD score is a discrete but ordered value, we report the percentage of cases whose classification lays within one class of the true value (\emph{one-off}). 
The Table \ref{table:1} summarizes the results of the experiments. 
Our method outperforms the state-of-the-art on predicting FEV1 and GOLD score. Adding the generative regularization ($\lambda_1 =10$) reduces the accuracy but results in much better interpretability. In the following, we study the effect of $\lambda_1$.

\textit{Generative regularizer ($\lambda_1$):} The Fig.~\ref{fig:results}(b) reports the spectral behaviors of the latent features (\ie $\phi_e(\mathcal{X}_i)$) for different values of $\lambda_1$. For small $\lambda_1$ the loss function doesn't optimize for the generative loss. Hence, the latent space representation becomes highly redundant, and all the attention weights $\alpha_{ij}$ becomes similar and converges to $\frac{1}{|\mathcal{X}_i|}$. The Fig.~\ref{fig:results}(c) shows the trade-off between effective rank  of the latent feature (red, $y$-axis on left) and $R^2$ for predicting FEV1 (blue, $y$-axis on right). Although, the $R^2$ drops a little, the rank, which represents the diversity of the latent features, improves drastically. The gap between accuracies of $\lambda_1=0$ and $\lambda_1>0$ is the price we pay for the interpretability. Fully generative model ($\lambda_1 \rightarrow \infty$) does not produce good prediction.

\begin{figure}
\centering
\subfigure[ ]{\label{fig:a}\includegraphics[height=34mm]{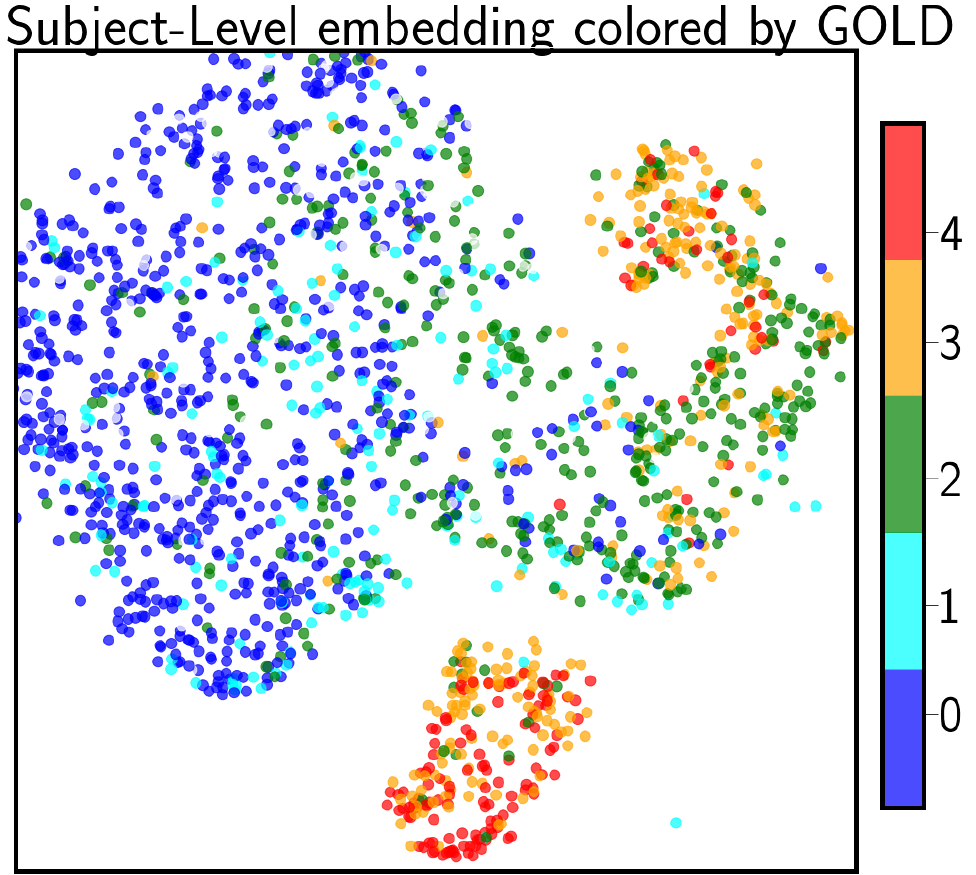}}
\subfigure[ ]{\label{fig:b}\includegraphics[height=34mm]{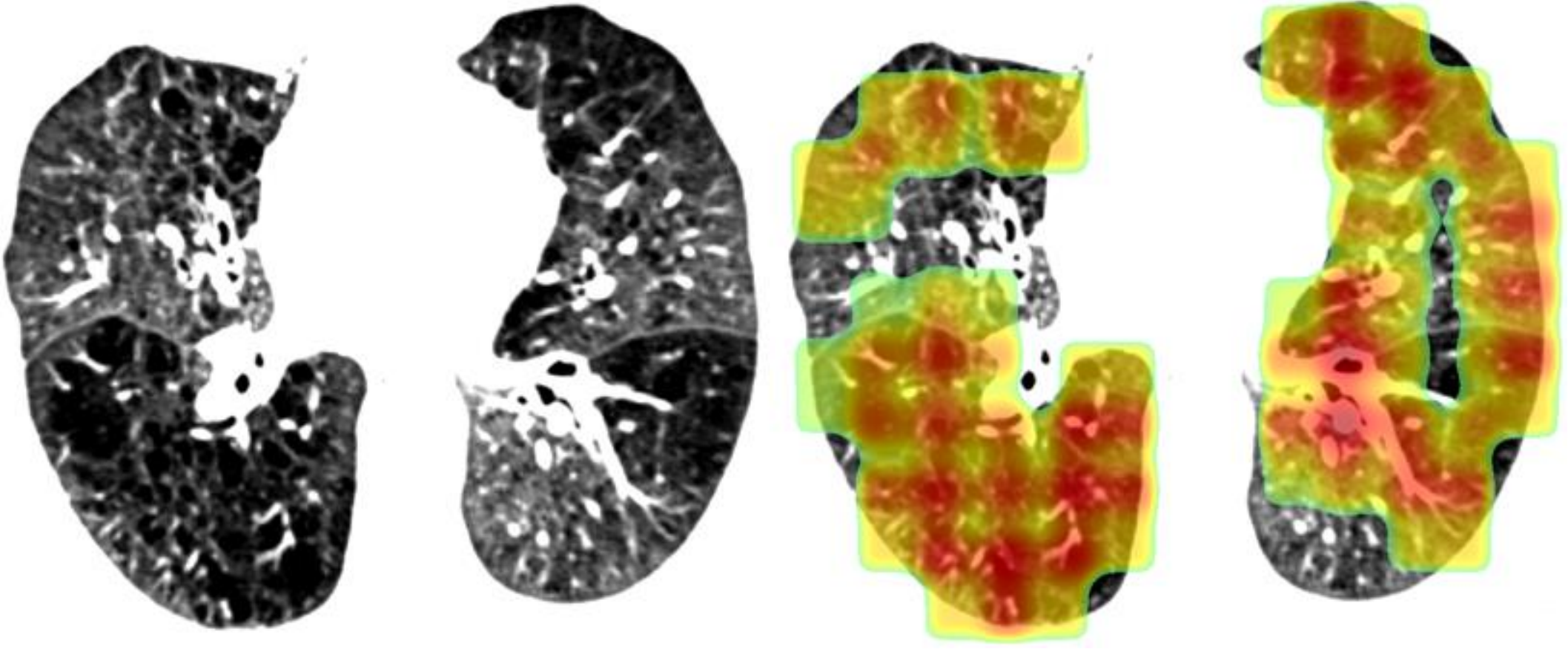}}
\caption{(a) Embedding the subjects in 2D using tSNE. The dots represents one subject colored by the GOLD score. (b) An axial view of the attention map on a subject. Red color indicate higher relevance to the disease severity.}
\label{fig:fig4}
\end{figure}

\textit{Visualization:} We use tSNE~\cite{Maaten2008VisualizingT-SNE} to visualize subject-level features in two dimension. In Fig.~\ref{fig:fig4}(a), each dot represents a subject colored by the GOLD score. Even in two dimension, subjects with GOLD score of (0,1) and (3,4) are quite separable and 2's are in between. The bimodal distribution of GOLD stages 3 and 4, is sensitive to t-SNE parameterization and requires further investigation.~\ref{fig:fig4}(b) visualizes the attention weights on one subject. The dark area on the left lung, which is severely damaged, received hight attention. \textcolor{red}{} 
\section{Conclusion}
We developed a novel attention-based model that achieves high prediction while maintaining interpretability. The method outperforms state-of-art and detects correct instances on the simulated data. Our current model does not account for spatial locations of the patches. As a future direction, we plan to extend the model to accommodate relationship between patches. 

\paragraph{Acknowledgement} This work is partially supported by NIH Award Number 1R01HL141813-01. We gratefully acknowledge the support of NVIDIA Corporation with the donation of the Titan X Pascal GPU. We thank Competitive Medical Research Fund (CMRF) grant for their funding.

\bibliography{Mendeley}
\bibliographystyle{splncs03}

\end{document}